# Puzzle Solving without Search or Human Knowledge: An Unnatural Language Approach

**David Noever**
PeopleTec, Inc.
Huntsville, AL
david.noever@peopletec.com

**Ryerson Burdick**
University of Maryland, College Park
College Park, MD
rburdick@terpmail.umd.edu

## ABSTRACT

The application of Generative Pre-trained Transformer (GPT-2) to learn text-archived game notation provides a model environment for exploring sparse reward gameplay. The transformer architecture proves amenable to training on solved text archives describing mazes, Rubik's Cube, and Sudoku solvers. The method benefits from fine-tuning the transformer architecture to visualize plausible strategies derived outside any guidance from human heuristics or domain expertise. The large search space ($>10^{19}$) for the games provides a puzzle environment in which the solution has few intermediate rewards and a final move that solves the challenge.

## INTRODUCTION

*Why transformers?* For natural language generation (NLG), the transformer architecture provides a scalable mechanism to encode long-range dependencies needed to output plausible text narratives. Transformers (Vaswani, et al., 2017) have rapidly advanced to rival or overtake other deep learning architectures such as convolutional neural networks (CNN). Initially developed to handle long-term language dependencies, this approach over-weights important relations via the "attention" method rather than attempting to localize dependencies (CNN) or grow dense networks for all weights. While the resulting sparse network extends available long-term connections needed to relate distant parts-of-speech or sentence context, the net effect has grown to massive models now in the trillions of connection weights (Child, et al., 2019). This approach has since found application in other fields unrelated to the original language modeling, such as non-local effects needed for visual context problems. Among the early successes, the Generative Pretrained Transformer (GPT-2) from Open AI (Radford, et al., 2019) remains one of the most robust architectures for fine-tuning applications. In these cases, the original training set gets specialized to diverse domains outside of its initial text data (Woolf, 2020). As a result, previous work has applied GPT-2 to play chess (Noever, et al., 2020), Go (Ciolino, et al., 2020), and other complex strategy games without knowing the explicit rules but instead learning the text patterns necessary to transfer learning from archival play. Since no move constraints get introduced to the transformer (e.g. legal vs. illegal moves), the trained model results in gameplay without human knowledge (Silver, 2017). Because of its origins in natural language modeling, GPT-2 serves as a viable mimic of human narratives (sometimes called a "stochastic parrot"), particularly for the specialized use case called here as "unnatural language" generation. Figure 1 highlights some example applications of learning text archives for puzzles including Rubik's cube, Sudoku, and maze solvers.

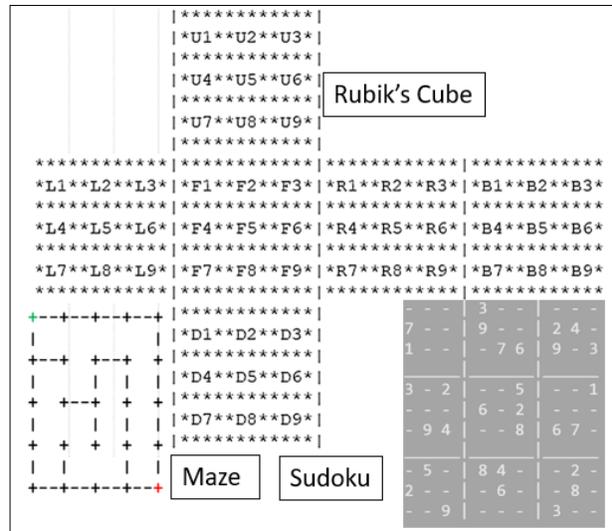

Figure 1. Example sparse reward puzzles in text notation

*Why puzzles and games?* The application of AI and machine learning approaches to gameplay offers a rich history ranging from Deep Blue in chess (1997) to AlphaZero (Silver, et al., 2017). One appealing aspect follows from the obvious scoring metrics associated with scoring humans vs. machines. In economics and game theory, a key distinction among the types of games amenable to AI implicitly favors perfect information games, such as chess, checkers, Go, etc. The board state is known equally to the human and machine players and gameplay progresses sequentially. The sequential play alternates its moves in a way different from simultaneous plays like Rock, Paper, Scissors, which are also perfect information but not alternating moves. Recent advances in Monte Carlo tree search (Silver, et al., 2017) have conquered human experts even in imperfect information games like poker, in which players can bluff while concealing their true game state until forced to reveal winners and losers in the final move of turning over cards or folding their hands. A third game category has recently attracted AI attention and might be informally classed as open-ended worlds like the video play in DOTA and StarCraft 2. Playing these games effectively as a tree search problem requires enormous computing resources and must handle the wide universe of available strategies ("where almost anything goes"). The present research examines a fourth possible category well known to the reinforcement learning community as games or puzzles that offer sparse rewards. These problems are generally characterized by large state spaces and a relatively small number of states which have an associated reward signal. Infrequent rewards often make gradient-based search and other methods that depend upon a smooth reward signal impractical.

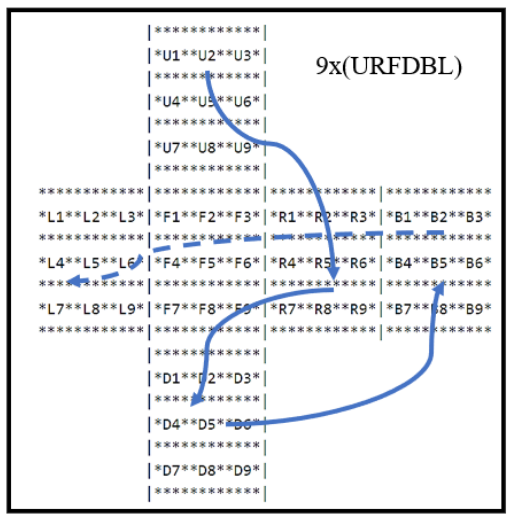

Figure 2. Rubik's Cube String Notation and Syntax for Position and Colors

One notable example of a sparse rewards task is the Rubik's cube. The Rubik's cube is a puzzle with 6 rotating faces, each composed of 9 smaller squares ("cubies") which take one of 6 colors. The objective is to rotate the faces until each face contains 9 squares of the same color. The Rubik's cube is an extreme example of a sparse rewards task (Demaine, et al., 2017; Darbandi, et al, 2020); it has a large state space consisting of approximately $4.9 \times 10^{19}$ possible configurations, and only the goal state has an associated reward signal (McAleer et al., 2018). This causes a sudden stepwise gain in rewards when making the final solving move.

A less extreme example of a sparse rewards task is the numerical puzzle game, Sudoku. The objective of Sudoku is to fill in missing cells of a 9×9 grid with the numbers 1-9, subject to the conditions that no number may appear twice in the same row, column, or 3×3 block. Because of these conditions, Sudoku is also known as a constraint satisfaction game. Like the Rubik's Cube, Sudoku has an enormously large state space, as there are approximately $6.671 \times 10^{21}$ valid Sudoku grids alone (Felgenhauer & Jarvis, 2006), and a reward signal is only achieved during the final step of the solving process.

It is worth noting that traditional Monte Carlo tree search techniques have exhaustive computing needs compared to GPT-2. For example, AlphaGo uses 1920 CPUs and 280GPUs (or $3000 in electricity costs) for each game (Rajput, 2021). The research explores solving these sparse reward games without reinforcement learning or Monte Carlo tree search. Instead, we apply the long-range rewards (weights)

found in current language transformers based on their attention strategies applied to text generators. The best-known examples of games with text generators largely focus on fine-tuning the GPT-2. Previous work has applied GPT-2 to perfect information games (e.g. chess, Go). For Sudoku and Rubik's Cube, deterministic (search) algorithms deliver sufficient quantities of good training data such that traditional deep learning techniques can solve the games using computer visions approaches and convolutional neural networks (Gaddam, et al., 2021; McAleer, et al., 2018). We propose to solve the games using text-based (ASCII) archives and fine-tune the transformer architecture to visualize another strategic solution to the sparse rewards challenges.

**METHODS**

*String Notation for Rubik's Cube Representations*. For the Rubik's task, we generated a dataset consisting of 5,000 pairs of initial cube configurations and corresponding solutions. To generate the initial configurations, a scrambling formula was created by randomly generating a sequence of moves to perturb the cube from the completed state. These scrambling formulas were anywhere between 1 and 5 moves in length, and an equal number of samples were generated for each possible scramble formula length. Once an initial configuration was determined, the cube state was represented by an encoding string following the format of Liu (2018). As illustrated in Figures 2-3, this encoding uses the cube string positions for an unfolded cube with ordered positions (9 digits) for the following faces: Up (U), Right (R), Front (F), Down (D), Back (B) and Left (L). The string order proves important (Kociemba, 2019) since a fully solved cube would have 9x(URFDBL) for the completed color faces. The position U1 can be any of the 6 standard colors (red, yellow, orange, blue, white, green). A starting state like "RBL…" means the right color (say, green) is in fixed position U1, the back color (say, red) is in position U2, etc. Finally, once all scrambling formulas were converted to encoding strings, duplicate cube states were removed from the dataset and the remaining samples were split into a training set containing 2404 samples and a test set containing 601 samples.

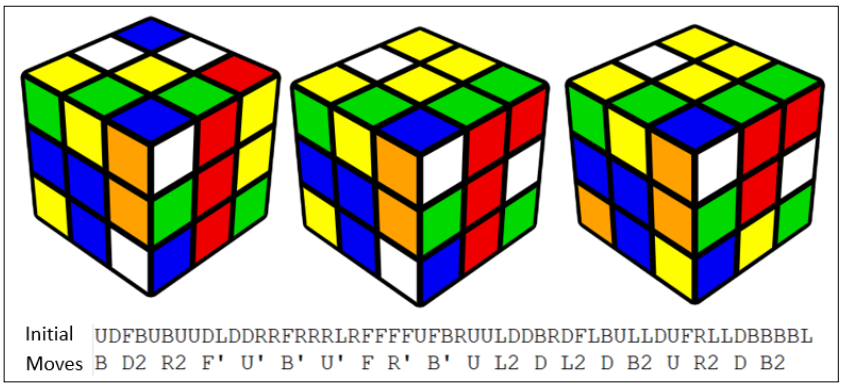

Figure 3. Solution Cube Notation for Visualizing Moves

After the initial Rubik's cube configurations and corresponding encodings were generated, a solution was determined using the Kociemba algorithm (Kociemba, 2019). The Rubik's solution syntax introduces each move as space-separated letters with punctuation and numbering conventions describing the turn. A single letter alone means to turn that face in the URFDBL dictionary of choices clockwise by 90 degrees (quarter turns). A letter with an apostrophe means the opposite counterclockwise turn by 90 degrees. If the letter has a number 2, the face gets a half-turn (180 degrees). An example initial state and solution of single moves is shown in Figure 3. We visualize each step of the cube solution using the Visual Cube application (Rider, 2017) and validate solutions using the PyCuber python library (Liaw, 2021).

*String Notation for Sudoku Representations*. For Sudoku, we collected one million solved games (Park, 2016), which consists of a similar split view of the initial and final state. To divide the start and finished puzzle, we insert a word prompt [WP] to demark the first digit of the 81 in the 9x9 puzzle (Figure 4). A zero value represents a blank or open slot. The second demarcation

Figure 4. Example Sudoku Starting and Final States

[RESPONSE] serves as a delimiter for the puzzle solution. The visualization of a solved puzzle was customized in a console application that pushes each new digit onto the string for replacing the next available open gap (zero). The puzzle's starting and ending delimiters (<|…|>) allow the generated text of a proposed solution to be parsed and truncated to simplify interpretation.

*String Notation for Maze Representations.* For solving mazes, we generated 10,000 random mazes and embedded their ASCII text solutions between the start and stop delimiters. To generate mazes of 4x4 and 5x5 (Rosettacode, 2021), we use (+) and (-) signs to outline the text grid boundaries, the use (|) pipe symbology to define walls. As shown in Figure 5, we encode both the unsolved and solved mazes in a single training text example for each maze. The training solutions follow the search methods outlined as breadth or depth-first techniques. (Sinck, 2021) Each example maze begins with the upper left corner as the starting position (**); the direction of maze navigation follows a text arrow notation (^^=up; >>=right; vv=down; <<=left). As with the other cases, the training set represents a series of maze pairings (unsolved and solved) with one maze in a single row submitted to the transformer.

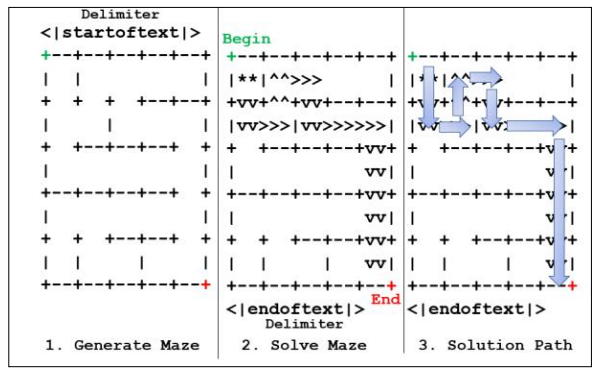

Figure 5. Maze generator and transformer solutions

## RESULTS

*Findings for Cube Solvers.* On the Rubik's Cube data, the transformer was unable to solve the complete puzzle more than one in seven attempts. Out of the 601 generated responses for the test examples, 11 were invalid (~1.8%), 576 were incorrect (~95.8%), and only 14 were correct (~2.3%). The small proportion of invalid generated responses indicates that despite being trained initially on natural language, the transformer has adapted well to the "unnatural" language of Rubik's cube formulae; even when it was unable to solve the cube, the overwhelming majority of the time the transformer produced an output which corresponds to a valid Rubik's formula. Figure 6 shows the solution for single rows as an incomplete solution but progressively improved cube state.

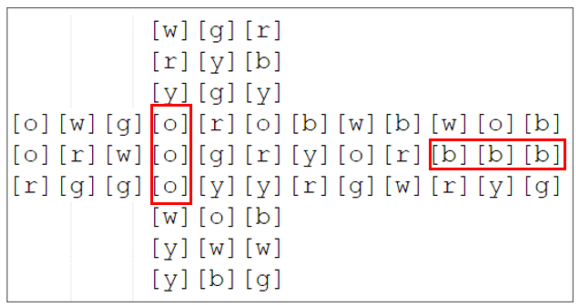

Figure 6. Rubik's Cube Transformer Solving for Single Rows

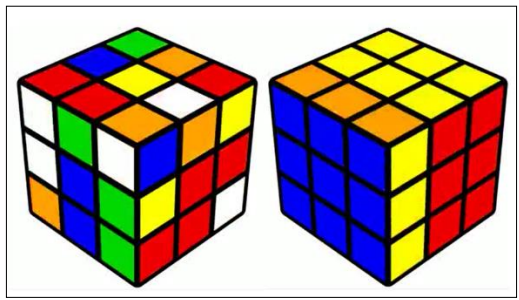

Figure 7. Transformer (left) vs. Kociemba (right) algorithm

Given the short fine-tuning period (~2000 epochs) and the small number of training examples (~2400), it is significant that the transformer was able to solve the Rubik's puzzle at all. Interestingly, though the majority (9/14) of correct generated responses were only 1-3 moves in length, the remaining correct responses were long: one response was 52 moves long, three were 53 moves long, and the longest was 61 moves. Given the small sample size, it is difficult to generalize about the transformer's performance. Regardless, the existence of these solutions suggests the transformer may have learned certain solving patterns present in the Kociemba algorithm.

A video comparing Rubik's Cube solutions is found online (Noever, 2021). Figure 7 compares the Kociemba algorithm (right) to the transformer solution (left) at the same time step. The algorithm solution shows a quarter turn before reaching the end with all six aligned colored faces after 71 steps. The transformer generates 64 steps before reaching the token limit (1024) for generated text outputs as an inherent GPT-2 limit. To illustrate the sparse rewards, neither the algorithmic nor transformer solution capitalizes on a partial reward, such as solving one color for a face or multiple faces in an intermediate step. The transformer did, however, occasionally solve for single rows and columns in instances where it was unable to solve the puzzle before reaching the token limit. An example of the Rubik's Cube transformer solving for rows and columns is shown in Figure 6.

*Findings for Sudoku Solvers.* Figure 8 shows the GPT-2 gameplay for Sudoku from a randomly selected initial state to a partial (but flawed) final solution. The orange diamonds show the repeated digits as errors in completing the square with unique numbers both in the interior square and the overall rows and columns. A validation algorithm that checks for repetitions [1-9] in every row, column, and sub-square could potentially serve as an overlay on generated text games, much in the same way that Chess game generators playing against humans filter out invalid moves. Because GPT-2 models include the training text formatting in their transformer architecture, the Sudoku training set may benefit from the native grid or matrix rather than string input which masks the sub-grid orientation. The resulting transformer would generate complete puzzle grids rather than require additional visualizations as shown in Figure 8 for a console (command-line) player.

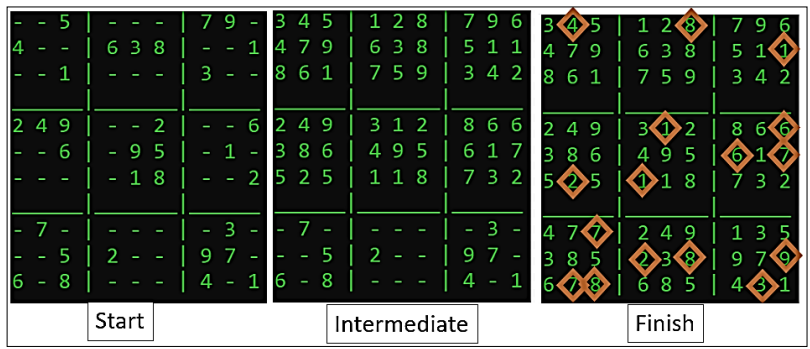

Figure 8. Sudoku Solution Stages using GPT-2

*Findings for Maze Solvers.* Figures 5 and 9 show transformer solutions to the 5x5 (Fig. 5) and 4x4 (Fig. 9) maze sizes. Unlike the Sudoku case, the maze training set preserves formatting for its basic maze grid without removing all end-of-line breaks as a single string. In this way, the maze resembles a narrative paragraph versus the Sudoku sentence format. The trained transformer outputs both a viable unsolved maze and its proposed solution as a pair bracketed by start and end delimiters. Since all outputs are generated unconditionally and without a prompt for a starting maze, the output appears as both a scenario generator (viable unsolved maze) and a solution generator (moves to complete the puzzle). Given the token limit of 1024 for generated text, the proposed maze sizes stop at 6x6 grids if the formatting is 4 spaces per grid as shown in Figure 9 and if the unconditional output includes both the starting maze and its paired solution. If a prompt or conditional model is run, the maze sizes naturally extend but the combinatorial moves limit the solution's viability.

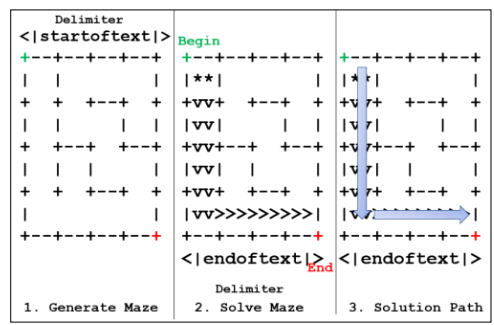

Figure 9. Transformer solution to text mazes in 4x4 size

**DISCUSSION**

Many other games with sparse reward signals have received attention from the reinforcement learning community, including Sokoban, Montezuma's Revenge, and Mountain Car (McAleer et al., 2018; Moore,

1990). Unlike these games, both Rubik's Cube and Sudoku are well-suited to the application of text generators because they conveniently allow for the examination of sparse rewards problems from within the confines of games with sequential play and discrete representations of state. Additionally, for both games, deterministic (search) algorithms can provide sufficient quantities of training data such that traditional deep learning techniques, e.g. CNNs, can solve them. Compared to denser reward games, the maze, Rubik's, and Sudoku puzzles require considerable exploration across a flat fitness or optimization landscape. In the case where a solution might take more computing resources to iterate exploratory steps, the attention mechanism behind GPT-2 offers a method to attack the contextual problem of knowing where the numbers or colored faces might relate to each other in the constrained volume of the cube or number squares. Figure 10 illustrates the Sudoku weights for layer 9 as an example of long-term attention and context between a starting number and its long-range dependencies. However, the transformer's ability to solve beyond the 1024 token limit of generated solutions limits the exploration to easier game starting points only. No transformer output for either game achieved a finished state from an arbitrarily random ("hard scrambled") state in the allotted number of steps. Instead, the transformer trained on nearly completed states (e.g. perturbed from a finished state) showed promise in accomplishing its goal to solve the puzzles. Just as with the chess and Go Transformers, the goal of generating plausible gameplay shows possible application but succeeds with supervision and filtering of illegal actions. The secondary goal of demonstrating rule-acquisition (plausible moves) suggests that explicit human knowledge of strategies or heuristics may not be needed specifically for opening or closing moves when the completion times fall within the attention limit of the transformer's context.

Well-known techniques in reinforcement learning emphasize turning a sparse reward game into a denser environment. These approaches feature human domain expertise to craft heuristics, such that the exploration space shrinks or partial rewards provide a stepping stone to reach the solution. A simple example would be solving a maze problem by recursive backtracking or applying the right-hand rule (Roberts, 2015). In the case of Rubik's Cube solvers, many intermediate steps might qualify as partial rewards, such as the layered method, cross, or daisy creations (Youcandothecube.com, 2021). As a bookkeeping strategy, human Sudoku solvers favor keeping track of which numbers are still possible for each square, thus iteratively narrowing the search space. The hard-coding of such heuristics however ranges outside the scope of the transformer architecture and its powerful capabilities to take raw text games as its only input without domain knowledge when fine-tuned to a new text source and format.

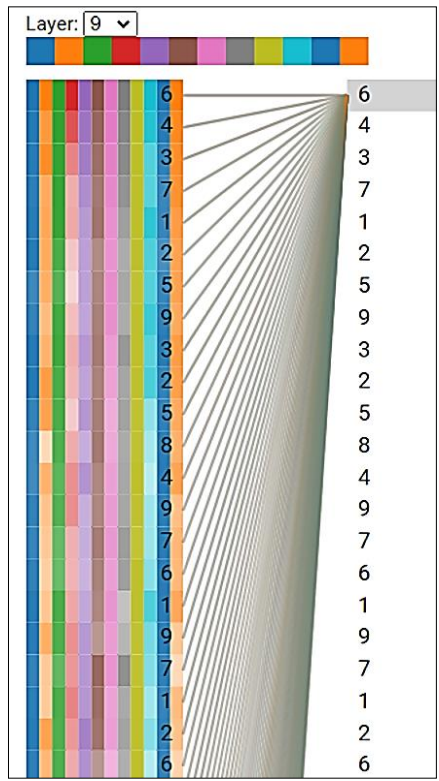

Figure 10. Layer visualization of long-range dependence for a single Sudoku game

One intriguing outcome of exploring transformers with sparse rewards is to suggest new approaches. The attention mechanism itself builds in overweighted connection strengths across longer-range context, a critical feature for language models. Ironically, one can posit that attention weights create a sparse reward landscape appropriate to generating interesting narrative text since a frequency-based word approach emphasizes common but less telling words (such as stop words "a", "the", etc.). In this way, attention-based models effectively balance the training dataset based on token interest and context rather than frequency. For games, the reinforcement learning community similarly maps flat gradient landscapes to maximize the ratio of rewarding exploitation steps compared to fruitless exploration ones. A simple strategy in sparse rewards substitutes "curiosity-driven" exploration, such that incremental rewards appear when going to points previously not visited. In Sudoku, one can imagine a similar exclusion priority or constraint geared towards not aimlessly substituting [1-9] digits when a row, column, or sub-square already has it. This approach prioritizes a restricted action. In the linguistic origins of GPT-2, the same reward or weight structure might favor novel word choices to avoid repetitive phrases.

The capability of transformers and other text generation methods to play games extends far beyond mazes, Rubik's Cube, and Sudoku. Previous research has highlighted their potential to generate plausible moves for other games which have historically served as benchmarks for game-playing algorithms, notably Chess (Noever et al., 2020) and Go (Ciolino et al., 2020). Other board games and puzzles offer additional angles from which to examine environments with sparse reward signals (Figure 11). Hex, a board game that has previously drawn attention from the AI community, is one such game. Like Rubik's and Sudoku, it is a perfect information game where the only obvious reward signal is triggered after the final, game-winning move. Unlike Rubik's and Sudoku, Hex is a competitive, 2-player game. It is also amenable to Smart Game Format (SGF), a common standardized notation for the textual representation of game states. Other candidate games and puzzles include TwixT, which is similar to Hex in both game layout and objective, and Tantrix, which offers sparse rewards in a competitive setting with more than 2 players.

| Text Game Format | Supported Games/ Puzzle with Possible Data |
|---|---|
| String Puzzle Notation | Rubiks Cube, Sudoku |
| Smart Game Format (SGF) | Amazons, Ataxx, Backgammon, Byte, Chase, Chess, Dvonn, Exxit, Focus, Gess, Gipf, Go, Gobblet, Gomoku+Renju, Hex, Hive, Hnefatafl, Jungle, Kropki, Kuba, Lines of Action, Neutron, Nine Men's Morris, Octi, Othello, Philosopher's Football, Phutball, Plateau, PÜNCT, Quadrature, Sahara, Shogi, Tamsk, Tantrix, Trax, Tripples, Tumbling Down, TwixT, Xiangqi, Yinsh, Zèrtz |
| Portable Game Notation (PGN) | Chess |
| Portable Draughts Notation | Checkers, Draughts |
| Bridge Notation | Bridge |
| Video Simulator Grammars | Multiple vintage games SimCity 2000, Pirates, Minecraft, ZZT, etc. |

Figure 11. Game & Puzzle Environments with Text Generation Play

**CONCLUSIONS**

Without encoding puzzle heuristics, the application of GPT-2 can generate viable moves in three sparse reward games: mazes, Rubik's Cube and Sudoku. These examples offer a novel text-based method to learn plausible moves without human instruction, heuristics, or explicit domain-specific rulesets. These puzzles provide appealing visualization environments to track algorithmic progress incrementally and score winning strategies, identify novel solutions, and augment the traditional black-box understanding inherent in large-scale transformers. Just as attention-based methods provide long-range context, future efforts for improving transformers in gameplay should emphasize larger token limits (>2048 in GPT-3) or condensed game notations for archives.

**ACKNOWLEDGEMENTS**

The authors would like to thank the PeopleTec Technical Fellows program and the Internship Program for encouragement and project assistance.